\pdfoutput=1

\documentclass[11pt]{article}

\usepackage{acl}

\usepackage{times}
\usepackage{latexsym}

\usepackage[T1]{fontenc}

\usepackage[utf8]{inputenc}

\usepackage{microtype}

\usepackage{graphicx}
\usepackage{amsmath}
\usepackage{pifont}%

\usepackage{booktabs}
\usepackage{xcolor}
\usepackage{float}
\usepackage{makecell}
\usepackage{multirow}
\usepackage[multiple]{footmisc}
\usepackage[normalem]{ulem}
\usepackage{xurl}
\usepackage{subcaption}
\usepackage{mwe}

\usepackage{enumitem,amssymb}
\newlist{todolist}{itemize}{2}
\setlist[todolist]{label=$\square$}
\usepackage{pifont}

\definecolor{darksalmon}{RGB}{230, 156, 116}
\definecolor{skyblue}{RGB}{115, 190, 229}

\def\modelname{XAMPLER\xspace}
\def\sib{SIB200\xspace}
\def\siblang{176\xspace}

\def\masakha{MasakhaNEWS\xspace}
\def\glot{Glot500\xspace}
\def\mala{MaLA500\xspace}

\usepackage{endnotes}

\newcounter{notecounter}

\newcommand{\enoteson}{\long\gdef\enote##1##2{{
			\stepcounter{notecounter}
			{\large\textbf{ \hspace{1cm}\arabic{notecounter} $<<<$ ##1: ##2 $>>>$\hspace{1cm}}}}}}
\enoteson

\title{\modelname: Learning to Retrieve Cross-Lingual In-Context Examples}

\author{Peiqin Lin$^{1,2}$, André F. T. Martins$^{3,4,5}$, Hinrich Schütze$^{1,2}$ \\
        $^1$Center for Information and Language Processing, LMU Munich \\
        $^2$Munich Center for Machine Learning \\
        $^3$Instituto Superior Técnico, Universidade de Lisboa (Lisbon ELLIS Unit) \\
        $^4$Instituto de Telecomunicações \quad
        $^5$Unbabel \\
        \texttt{linpq@cis.lmu.de}
}

\begin{document}
\maketitle
\begin{abstract}
Recent studies indicate that leveraging off-the-shelf or fine-tuned retrievers, capable of retrieving relevant in-context examples tailored to the input query, enhances few-shot in-context learning for English.
However, adapting these methods to other languages, especially low-resource ones, poses challenges due to the scarcity of cross-lingual retrievers and annotated data.
Thus, we introduce \textbf{\modelname: Cross-Lingual Example Retrieval}, a method tailored to tackle the challenge of cross-lingual in-context learning \textbf{using only annotated English data}.
\modelname first trains a retriever based on \glot, a multilingual small language model, using positive and negative English examples constructed from the predictions of a multilingual large language model, i.e., \mala.
Leveraging the cross-lingual capacity of the retriever, it can directly retrieve English examples as few-shot examples for in-context learning of target languages.
Experiments on two multilingual text classification benchmarks, namely \sib with \siblang languages and \masakha with 16 languages, demonstrate that \modelname substantially improves the in-context learning performance across languages.
Our code is available at \url{https://github.com/cisnlp/XAMPLER}.
\end{abstract}

\section{Introduction}

Large language models (LLMs) have shown emergent abilities in in-context learning, where a few input-output examples are provided with the input query. Through in-context learning, LLMs can yield promising results without any parameter updates \citep{DBLP:conf/nips/BrownMRSKDNSSAA20}. However, the efficacy of in-context learning is highly dependent on the selection of the few-shot examples \citep{DBLP:conf/acl-deelio/LiuSZDCC22}.

Recent studies \citep{DBLP:journals/corr/abs-2401-11624} have uncovered a more strategic approach to example retrieval. Rather than relying on random selection, these studies advocate for retrieving examples tailored to the input query, resulting in notable performance enhancements in in-context learning. The retrievers employed by these methods can be categorized into two main types: general off-the-shelf retrievers \citep{DBLP:conf/acl-deelio/LiuSZDCC22}, e.g., Sentence-BERT \citep{DBLP:conf/emnlp/ReimersG19}, and task-specific fine-tuned retrievers \citep{DBLP:conf/naacl/RubinHB22}, which are trained based on LLM signals (whether an example is helpful) using labeled data.

\begin{figure}
  \centering
  \resizebox {\columnwidth} {!} {
    \includegraphics[clip, trim=8.25cm 5.5cm 9.25cm 3.5cm]{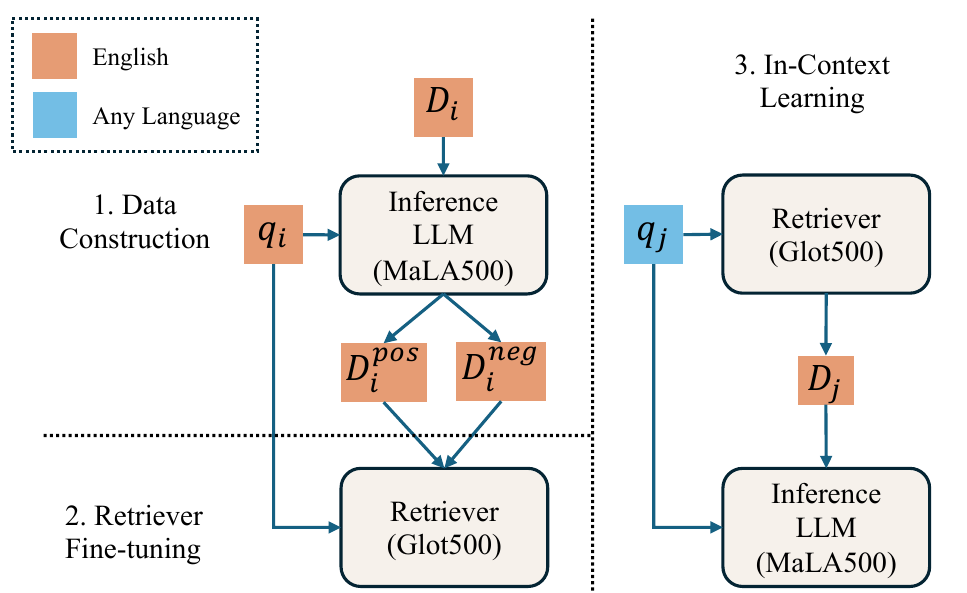}
  }
  \caption{\modelname involves three steps: 1. Data
  Construction: given a query in English $q_i$, we divide
  the candidate English examples $D_i$ into positive examples $D_{i}^{pos}$ and negative examples $D_{i}^{neg}$ based on the prediction of
  \mala \citep{DBLP:journals/corr/abs-2401-13303};
  2. Retriever Fine-tuning: we fine-tune the retriever based
  on \glot \citep{DBLP:conf/acl/ImaniLKSSKMSMYS23} using the
  constructed data; 3. In-Context Learning: given a query in
  any language $q_j$, we use the fine-tuned retriever to
  retrieve relevant English examples $D_j$ as few-shots for in-context learning.
  \textbf{%
  For training, \modelname requires English data only. Once trained, the model can be applied to any of the 500 languages covered by MaLA500/Glot500 without any need for (often unavailable) labeled low-resource data.}
  }
  \label{fig:overview}
\end{figure}

Utilizing off-the-shelf retrievers has been further validated as an effective approach in multilingual settings \citep{DBLP:journals/corr/abs-2212-09651,DBLP:journals/corr/abs-2306-10964,DBLP:conf/acl/TanwarDB023}.
However, this method encounters limitations when applied to low-resource languages.  Existing multilingual retrievers, e.g., SBERT \citep{DBLP:conf/emnlp/ReimersG20}, cover a limited number of languages (i.e., 50+), and language-model-based retrievers \citep{DBLP:conf/icml/HuRSNFJ20}
struggle to effectively align distant languages  \citep{DBLP:conf/iclr/CaoKK20,DBLP:journals/corr/abs-2311-08849}. Additionally, relying on off-the-shelf retrievers might lead to sub-optimal performance. Conversely, adopting task-specific fine-tuned retrievers has been demonstrated as a more effective approach \citep{DBLP:conf/naacl/RubinHB22}. Nonetheless, the availability of data for fine-tuning task-specific retrievers in low-resource languages is limited.

To tackle these challenges, we propose a simple yet effective method that relies solely on annotated English data, termed \modelname (Cross-Lingual Example Retrieval). As shown in Fig.~\ref{fig:overview}, given an English query $q_i$ and an English example from the candidate pool $D_i$, we employ in-context learning with \mala \citep{DBLP:journals/corr/abs-2401-13303}, a 10B multilingual LLM covering 534 languages, to predict the label of the query. Based on the correctness of the prediction, we classify the candidate example as either positive or negative, i.e., $D_{i}^{pos}$ and $D_{i}^{neg}$. Then, leveraging the curated dataset, we train a retriever based on \glot \citep{DBLP:conf/acl/ImaniLKSSKMSMYS23}, a multilingual small language model covering 534 languages, aiming to minimize the contrastive loss \citep{DBLP:conf/naacl/RubinHB22,DBLP:conf/emnlp/ChengHBZLW0WDZ23,DBLP:journals/corr/abs-2305-14128}. Finally, the trained retriever is directly applied to retrieve valuable few-shot examples in English for the given query in the target language. The retrieved English few-shot examples, along with the input query, are then fed into \mala for in-context learning.
Experiments across \siblang languages on \sib and 16 languages on \masakha show that \modelname effectively retrieves cross-lingual examples, thereby enhancing in-context learning across languages. %

\section{Approach}

\subsection{Problem Definition}

Given an input query $q_i$ in any language, our objective is to enhance in-context learning for predicting the label of $q_i$ by retrieving tailored few-shot examples from the pool of candidate examples $D$. Due to the scarcity of annotated data in low-resource languages, we introduce \modelname, namely, Cross-Lingual Example Retrieval. On one hand, we leverage in-domain English examples as the pool of candidate examples $D$, from which we retrieve cross-lingual examples in English for $q_i$ in any target language.
On the other hand, we only consider $q_i$ sourced from English training data to train the task-specific retriever, which is then directly applied for evaluation across languages. %

\subsection{Data Construction}

To train the task-specific retriever aimed at retrieving informative examples for the given query $q_i$, we consider contrastive learning, which requires both positive and negative examples for each query $q_i$.
We define examples as positive when the LLM accurately predicts the ground truth of $q_i$ while utilizing the example as a one-shot example appended to $q_i$ for in-context learning. Conversely, examples are categorized as negative if the LLM's prediction deviates from the ground truth.

Scoring all pairs of training examples presents a quadratic
complexity in $|D|$, making it resource-intensive. Inspired
by \citet{DBLP:conf/naacl/RubinHB22}, we mitigate this by
selecting the top $k$ similar examples as candidates. We
utilize Sentence-BERT
(SBERT) \citep{DBLP:conf/emnlp/ReimersG20}\footnote{We use
version distiluse-base-multilingual-cased-v1.} for candidate selection. Based on our experiments detailed in Section \ref{sec:effect_k}, we set $k=10$.
The top $k$ candidates for $q_i$ are denoted as $D_{i} = \{d_{i,1}, \cdots, d_{i,k}\}$, where each candidate $d_{i,j}$ is represented as $(x_{i,j}, y_{i,j})$, with $x_{i,j}$ being the input and $y_{i,j}$ the corresponding label.

After obtaining the candidate-query pairs $\{(q_i, d_{i,1}), \cdots, (q_i, d_{i,k})\}$, we conduct 1-shot in-context learning with \mala \citep{DBLP:journals/corr/abs-2401-13303} to predict the class of the $q_i$ given the candidate $d_{i, j}$, resulting in a predicted label  $\hat{y}_{i,j}$. %
If \mala correctly predicts the label of $q_i$ (i.e., $\hat{y}_{i,j}=y_i$), we consider the candidate $d_{i,j}$ as a positive example ($d^+_{i,j}$); otherwise
a negative example ($d^-_{i,j}$). Finally, we divide $D_i$ into sets of positive and negative examples, denoted as $D_i^{pos}$ and $D_i^{neg}$, respectively.

\subsection{Retriever Fine-tuning}

We utilize the contrastive loss \citep{DBLP:conf/naacl/RubinHB22,DBLP:conf/emnlp/ChengHBZLW0WDZ23,DBLP:journals/corr/abs-2305-14128} to train the task-specific retriever, aiming to maximize the similarity between $q_i$ and $x_{i,j}$ if $x_{i,j}$ is a positive example while minimizing the similarity
if $x_{i,j}$ is a negative example.
We opt for \glot \citep{DBLP:conf/acl/ImaniLKSSKMSMYS23} with a model size of 395M as the base model for training the retriever, considering the significant cost of fine-tuning an LLM. We train for 50 epochs using the AdamW optimizer with a learning rate of 2e-5 and a batch size of 16. Due to the multilingual nature of \glot, the fine-tuned retriever can be effectively transferred to retrieve in-context examples for other languages. %

\subsection{In-Context Learning}

At test time, when employing in-context learning across languages, where $q_i$ can be in any language, we use the fine-tuned task-specific retriever to retrieve a few cross-lingual examples in English tailored to $q_i$. The retrieved examples are appended to $q_i$ as input for \mala \citep{DBLP:journals/corr/abs-2401-13303} to predict the label of $q_i$ through in-context learning.

\section{Experiment}

\subsection{Setup}

\paragraph{Benchmark} We evaluate \modelname on two text classification benchmarks: \sib \citep{DBLP:journals/corr/abs-2309-07445} and \masakha \citep{DBLP:conf/ijcnlp/AdelaniMAATMODONEASDNMANOOOMMASYGABA23}. The partitioning for these datasets was predefined by the respective benchmarks. \sib is a massively multilingual text classification benchmark with seven classes. %
Our evaluation spans a diverse set of \siblang languages, obtained by intersecting the language sets of \sib and \mala (see \S\ref{sec:detailed_results}). The English training set contains 701 samples, with 204 evaluation samples per language. \masakha is a news
classification task for 16
African languages, covering six topics. %
The English training set contains 3.31k samples, with 175 to 948 evaluation samples per language.

Our evaluation framework follows the prompt template used in \citet{DBLP:journals/corr/abs-2401-13303}: `The topic of the news [sentence] is [label]', where [sentence] represents the text for classification and [label] is the ground truth. [label] is included when the sample serves as a few-shot example but is omitted when predicting the sample. We opt for English prompt templates over in-language ones due to the labor-intensive nature of crafting templates for non-English languages, especially those with limited resources. \mala takes the concatenation of few-shot examples and $q_i$ as input, then proceeds to estimate the probability distribution across the label set. We measure the performance with accuracy.

\paragraph{Baselines}
We compare \modelname with the following retrieving strategies:

\textit{Random Sampling.} We randomly select examples from the English candidate pool $D$.

\textit{Multilingual Language Models.} We use two massively multilingual language models, \glot and \mala, as retrievers.
Tailored examples are retrieved based on the cosine similarity between the sentence representations of the candidate and the query.
For \glot, we utilize mean pooling over hidden states of the selected layer. For \mala, we adopt a position-weighted mean pooling method on the selected layer, assigning higher weights to later tokens \citep{DBLP:journals/corr/abs-2202-08904}. We use K-Nearest Neighbors to select the layer that performs best across layers (see \S\ref{sec:layer}).
The selected layers for \glot and \mala are 11 and 21, respectively.

\textit{Off-the-shelf Retriever.} We also employ three off-the-shelf retrievers trained on parallel data: SBERT \citep{DBLP:conf/emnlp/ReimersG19}, LaBSE \citep{DBLP:conf/acl/FengYCA022}, and Multilingual E5 \citep{DBLP:journals/corr/abs-2402-05672}.\footnote{\url{https://huggingface.co/intfloat/multilingual-e5-large}}

We set the number of shots as the number of classes and evaluate under two settings: the label-aware setting, where one shot is provided per class, and the label-agnostic setting, where the most similar examples are retrieved regardless of their labels.

\begin{table}
    \centering 
    \resizebox{\columnwidth}{!}{
    \begin{tabular}{c|cc|cc}
    \toprule
    & \multicolumn{2}{c|}{\sib} & \multicolumn{2}{c}{\masakha} \\
    & Label-Aware & Label-Agnostic & Label-Aware & Label-Agnostic \\
    \midrule
    Random & 65.24 & 61.68 & 72.32 & 72.39 \\
    \glot & 66.60 & 68.55 & 73.35 & 73.01 \\
    \mala & 66.75 & 66.25 & 73.39 & 71.58 \\
    SBERT & 67.13 & 66.59 & 73.24 & 72.8 \\
    LaBSE & 68.51 & 73.69 & 72.54 & 73.29 \\
    Multilingual E5 & 69.09 & 74.61 & 73.63 & 72.61 \\
    \midrule
    \modelname & \textbf{70.18} & \textbf{75.91} & \textbf{75.02} & \textbf{73.85} \\
    \bottomrule
    \end{tabular}
    }
    \caption{Average macro-accuracy across the evaluated languages on \sib and \masakha using \modelname compared to the baselines. 
    }
    \label{tab:main}
\end{table}

\subsection{Main Results}

The comparison between the baselines and \modelname is illustrated in Table~\ref{tab:main}.
Our analysis reveals several insights based on the performance with In-Context Learning (ICL) across different methods. Notably, the random baseline exhibits the worst performance among the baselines using ICL,
emphasizing the critical role of example selection for effective in-context learning.
Multilingual language models, which retrieve examples based on semantic similarity learned during pretraining,
slightly outperform the random baseline. Leveraging an off-the-shelf retriever further improves performance, with Multilingual E5 emerging as the top performer among them.
Among all the methods, \modelname
achieves the highest performance in in-context learning. Specifically, on \sib, \modelname surpasses Multilingual E5 by 1.09\% in the label-aware setting and 1.30\% in the label-agnostic setting, and by 1.93\% and 1.24\% on \masakha, respectively.

Our two established practices to reduce resource requirements, i.e., selecting the top-k similar examples as candidates using existing off-the-shelf retrievers and using a smaller base model for retriever training, ensures that XAMPLER operates efficiently. For example, on the SIB200 benchmark (701 samples, 10 candidates per sample) using an NVIDIA GeForce GTX 1080 Ti (11GB), retriever fine-tuning took just 1.5 hours, and retrieving similar examples added only 0.1 seconds per query during inference. 

\subsection{Ablation Study}

To further assess the effectiveness of \modelname, we conduct the following ablation studies:

\begin{table}[ht]
    \centering
    \small
    \begin{tabular}{c|c}
    \toprule
    Method & Accuracy (\%) \\
    \midrule
    XLT (\glot) & 69.51 \\
    XLT (\mala) & 69.90 \\
    MT & 74.50 \\
    KNN & 72.85 \\
    \midrule
    \modelname & \textbf{75.91} \\
    \bottomrule
    \end{tabular}
    \caption{Performance comparison of \modelname with ablation methods on the \sib benchmark.}
    \label{tab:ablation}
\end{table}

\textit{Cross-lingual Transfer (XLT).} Cross-lingual transfer is another approach that utilizes English data. In this method, the multilingual language model is fine-tuned on English data and then evaluated across target languages. Both \glot and \mala are included, with their respective cross-lingual transfer methods denoted as XLT (\glot) and XLT (\mala). For \glot, we perform full-parameter fine-tuning with a batch size of 16 and a learning rate of 1e-5. For \mala, which is trained by incorporating LoRA \citep{DBLP:conf/iclr/HuSWALWWC22} into LLaMA 2-7B \citep{DBLP:journals/corr/abs-2307-09288}, we update only the LoRA parameters with prompt tuning. The learning rate is set to 1e-3, weight decay is 0.1, the maximum sequence length is 128, and the batch size is 16. The optimizer used is AdamW.

\textit{Machine Translation (MT).} We translate the examples retrieved by \modelname from English to the target language and use these translated examples as few-shot examples for in-context learning. For translation, we use the distilled 600M variant of NLLB-200 \citep{DBLP:journals/corr/abs-2207-04672}.

\textit{K-Nearest Neighbors (KNN).} We consider KNN
with the fine-tuned task-specific retriever of \modelname for comparison. Specifically, we adopt majority voting based on the labels of the examples retrieved by the given retriever.

\begin{figure}[t]
  \centering
  \resizebox {\columnwidth} {!} {
    \includegraphics[clip]{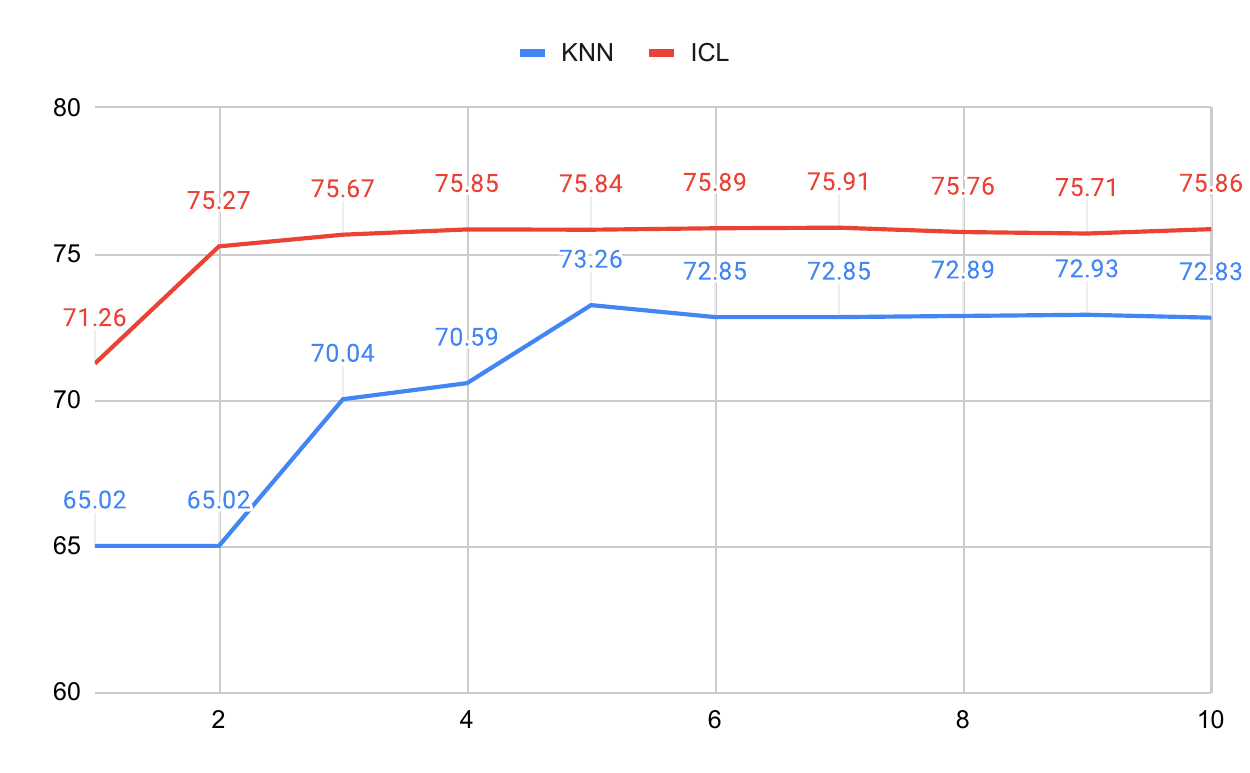}
  }
  \caption{KNN (K-Nearest Neighbors) vs. ICL (In-Context Learning) with different number of shots. X-axis: number of shots. Y-axis: Macro-average accuracy.}
  \label{fig:knn_vs_icl}
\end{figure}

The results of the ablation studies are shown in Table~\ref{tab:ablation}. 
\modelname outperforms XLT (\glot) and XLT (\mala) by 6.40\% and 6.01\%, respectively, demonstrating that in-context learning, when combined with informative cross-lingual few-shot examples, is superior to traditional cross-lingual transfer methods via fine-tuning.
Translating English examples into target languages results in slightly worse results than \modelname by 1.41\%. It demonstrates that \modelname benefits more from English in-context examples than in-language examples. A comparison between \modelname and KNN shows that \modelname performs better by 3.06\%. We also compared \modelname's performance with KNN and ICL using varying numbers of retrieved examples, as illustrated in Figure~\ref{fig:knn_vs_icl}. Interestingly, \modelname with ICL exhibits inconsistent superiority over KNN, with performance variances ranging from 3\% to 10\%. Specifically, \modelname with KNN achieves its peak performance with 5 examples, whereas ICL achieves impressive results with only 2 examples. Notably, in comparison to KNN's optimal performance, recorded at 73.26\% with 5 shots, \modelname with ICL demonstrates a notable improvement of 2.58\%. These findings underscore the efficacy of applying in-context learning in effectively leveraging the retrieved examples.

\section{Related Work}

Early studies \citep{DBLP:conf/acl/GaoFC20,DBLP:conf/acl-deelio/LiuSZDCC22,DBLP:conf/naacl/RubinHB22} on retrieving informative examples for few-shot in-context learning often rely on off-the-shelf retrievers to gather semantically similar examples to the query.

While off-the-shelf retrievers have shown promise, the examples they retrieve may not always represent optimal solutions for the given task, potentially resulting in sub-optimal performance. Hence, \citet{DBLP:conf/naacl/RubinHB22} delve into learning-based approaches: if an LLM finds an example useful, the retriever should be encouraged to retrieve it. This approach enables direct training of the retriever using signals derived from query and example pairs in the task of interest.

Several works \citep{DBLP:journals/corr/abs-2109-07684,DBLP:conf/emnlp/0010Z0L22,DBLP:conf/ijcnlp/WinataWKSP22,DBLP:journals/corr/abs-2212-09651,DBLP:journals/corr/abs-2306-10964,DBLP:conf/acl/TanwarDB023,DBLP:conf/naacl/CahyawijayaLF24} extend these methods to non-English languages. A study closely related to ours is \citet{DBLP:conf/emnlp/0010Z0L22}, which trains a cross-lingual example retriever via distilling the LLM’s scoring function and evaluates it on four languages for the Text-to-SQL Semantic Parsing task. However, our contribution lies in addressing the more challenging low-resource scenario, thereby extending the applicability and robustness of the approach proposed by \citet{DBLP:conf/emnlp/0010Z0L22}.

\section{Conclusion}

In this paper, we introduce \modelname, a novel approach designed for cross-lingual example retrieval to facilitate in-context learning in any language. Relying solely on English data, \modelname trains a task-specific retriever capable of retrieving cross-lingual English examples tailored to any language query, thereby facilitating few-shot in-context learning for any language. Experiments on \sib
and \masakha
show that \modelname outperforms previous methods by a notable margin.

\section*{Limitations}

We did not consider other models and benchmarks due to the absence / unavailability of massively multilingual ones. Additionally, while it is acknowledged that English may not universally serve as the optimal source language for cross-lingual transfer across all target languages \citep{DBLP:conf/acl/LinCLLZXRHZMALN19,DBLP:journals/corr/abs-2305-00090,DBLP:conf/eacl/LinHZMS24}, our study does not explore the selection of different source languages due to the predominant availability of training data in English for many tasks. 

\section*{Acknowledgements}
This work was funded by DFG (SCHU 2246/14-1),
the European Research Council (DECOLLAGE, ERC-2022-CoG \#101088763),  EU's Horizon Europe Research and Innovation Actions (UTTER, contract 101070631), by the Portuguese Recovery and Resilience Plan through project C645008882-00000055 (Center for Responsible AI),  by the DAAD programme Konrad Zuse Schools of Excellence in Artificial Intelligence, sponsored by the Federal Ministry of Education and Research, 
and by FCT/MECI through national funds and when applicable co-funded EU funds under UID/50008: Instituto de Telecomunicações. 
\bibliography{anthology,custom}

\clearpage
\appendix
\section{KNN Performance Across Layers}
\label{sec:layer}

\begin{figure}[ht]
  \centering
  \resizebox {\columnwidth} {!} {
    \includegraphics[clip]{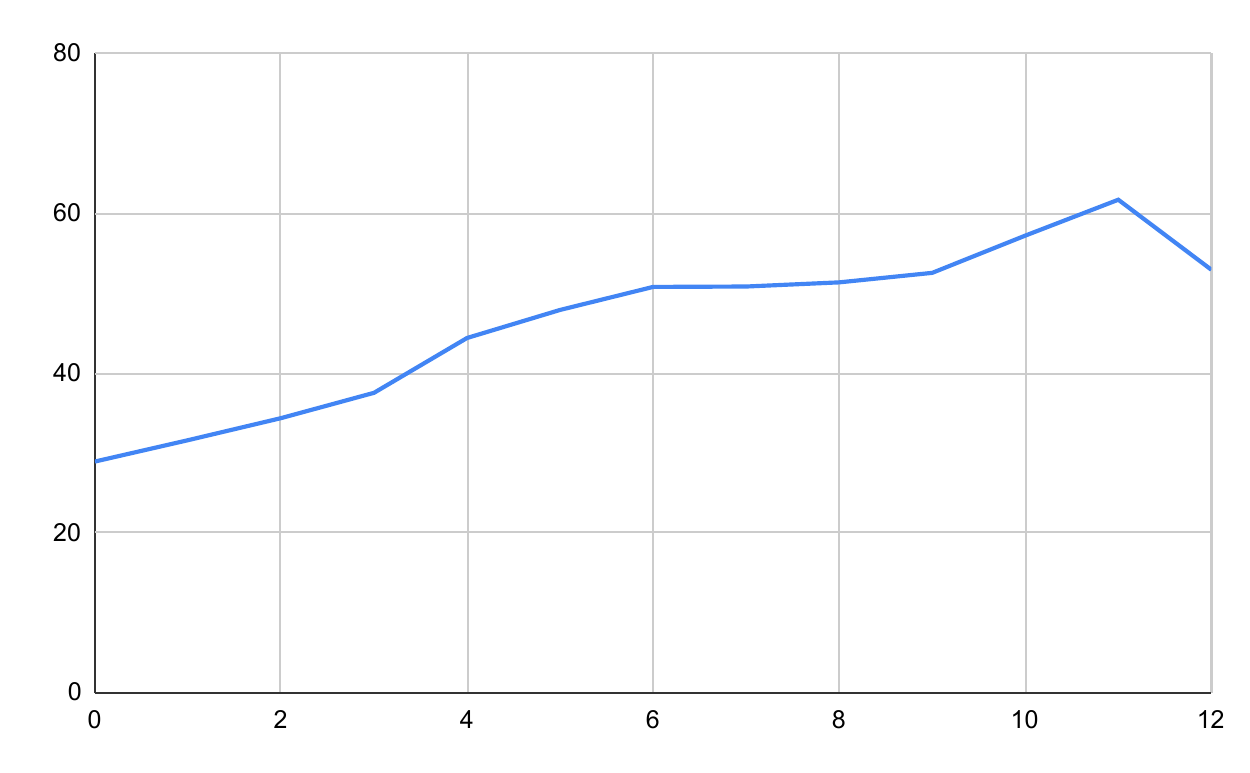}
  }
  \caption{Results of 10-shot KNN (K-Nearest Neighbors) with \glot as retriever across layers.}
  \label{fig:glot_layer}
\end{figure}

\begin{figure}[ht]
  \centering
  \resizebox {\columnwidth} {!} {
    \includegraphics[clip]{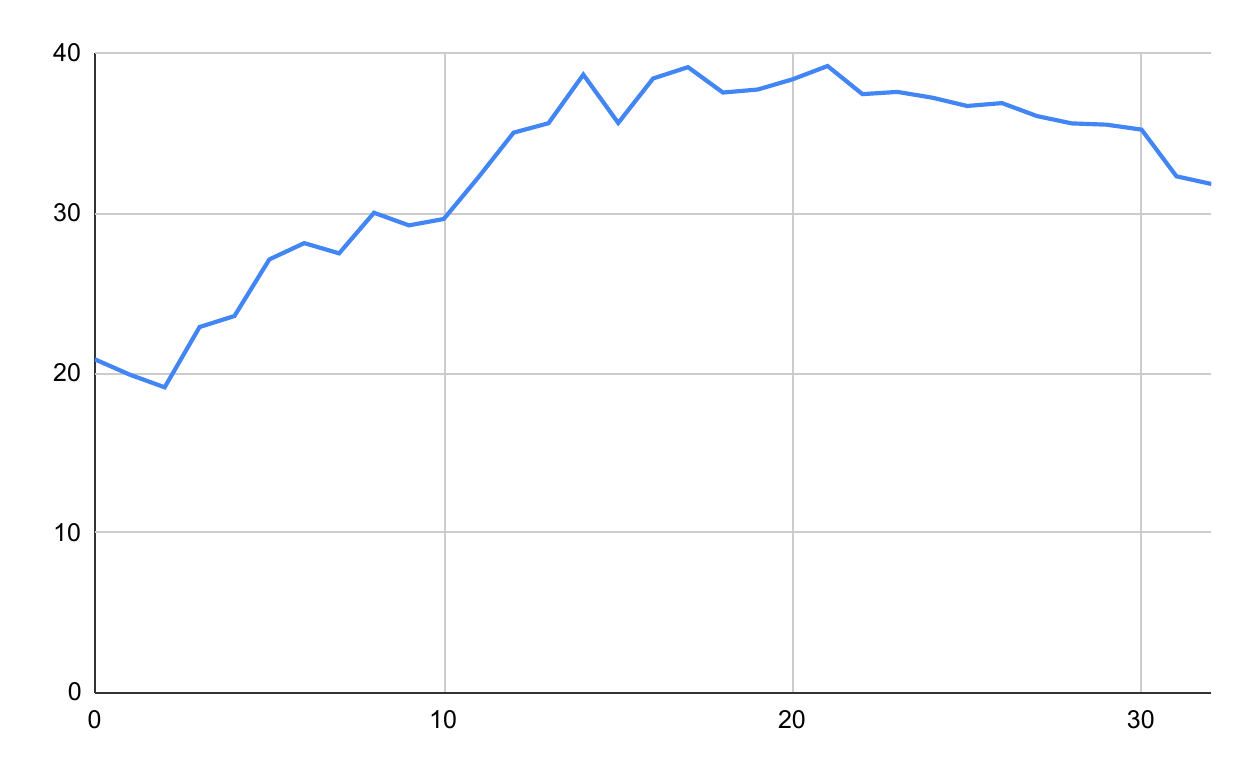}
  }
  \caption{Results of 10-shot KNN (K-Nearest Neighbors) with \mala as retriever across layers.}
  \label{fig:mala_layer}
\end{figure}

We show the 10-shot KNN results across layers with \glot and \mala as retrievers in Figure~\ref{fig:glot_layer} and \ref{fig:mala_layer}. As shown, layer 21 of \mala and layer 11 of \glot achieve the best performance across layers. Therefore, the retrieved results based on these two layers are used in the baselines.

\section{Effect of $k$}
\label{sec:effect_k}

\begin{figure}[ht]
  \centering
  \resizebox {\columnwidth} {!} {
    \includegraphics[clip]{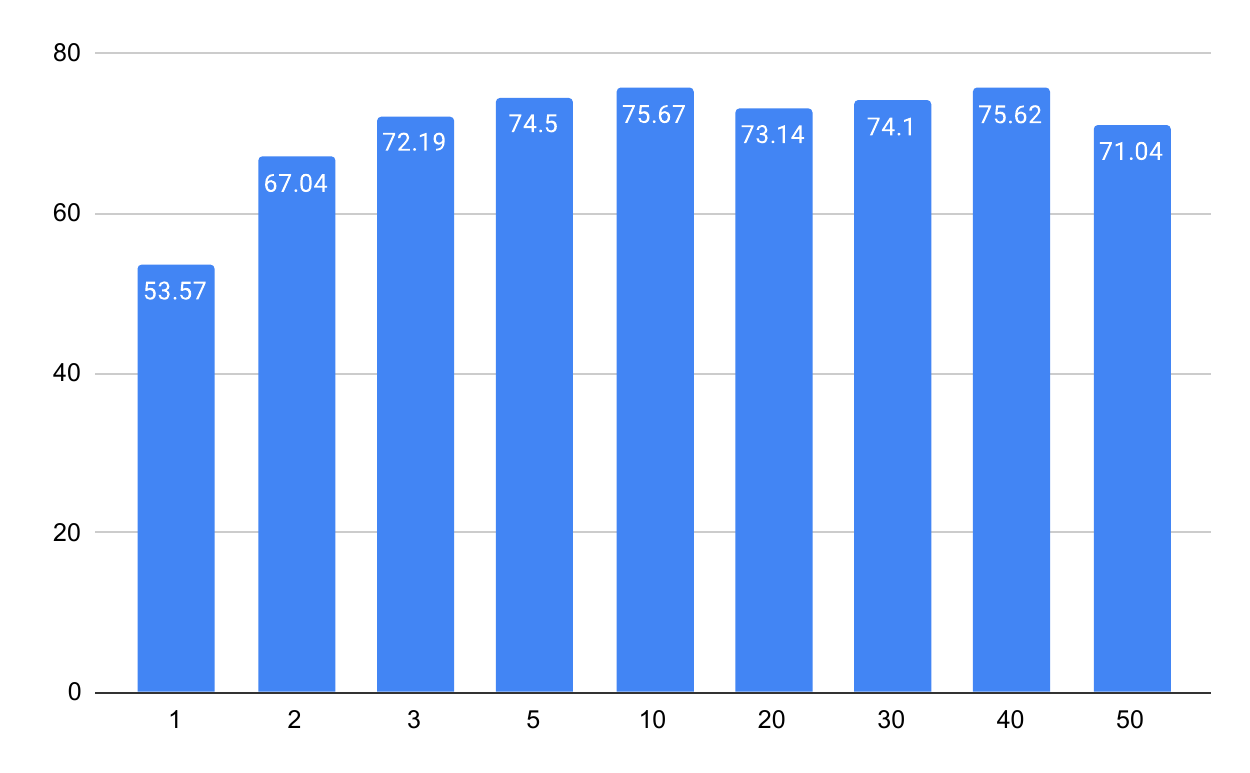}
  }
  \caption{In-context learning with \modelname with different $k$.}
  \label{fig:effect_k}
\end{figure}

We conduct additional experiments to analyze the impact of the parameter $k$, with the results presented in Figure~\ref{fig:effect_k}. Our findings indicate that \modelname performs optimally when $k=10$. However, as $k$ exceeds 10, there is a slight decrease in performance. This trend may be attributed to the possibility that increasing $k$ leads to fewer hard negatives for training the retriever.

\section{Detailed Results}
\label{sec:detailed_results}

The language list of \sib and the results of \modelname and the compared baselines are shown in Table~\ref{result1} and Table~\ref{result2}. The language list of \masakha and the results of \modelname and the compared baselines are shown in Table~\ref{result}.

\begin{table*}[ht]
    \centering
    \resizebox{0.8\textwidth}{!}{
\begin{tabular}{c|ccccccc|ccccccc}
\toprule
& \multicolumn{7}{c|}{Label-Aware} & \multicolumn{7}{c}{Label-Agnostic} \\
& Random & \glot & \mala & SBERT & LaBSE & Multilingual E5 & \modelname & Random & \glot & \mala & SBERT & LaBSE & Multilingual E5 & \modelname \\
\midrule
ace\_Latn & 72.55 & 75.00 & 72.06 & 74.51 & 72.55 & 74.51 & 72.06 & 63.73 & 71.57 & 70.10 & 70.10 & 76.47 & 80.39 & 76.96 \\
acm\_Arab & 62.25 & 64.71 & 65.20 & 70.10 & 66.18 & 68.63 & 75.00 & 61.76 & 71.08 & 65.69 & 78.92 & 75.98 & 75.98 & 79.90 \\
afr\_Latn & 77.45 & 76.47 & 77.94 & 79.90 & 80.39 & 81.86 & 79.41 & 70.59 & 79.41 & 78.43 & 83.82 & 88.24 & 81.86 & 87.25 \\
ajp\_Arab & 64.71 & 67.65 & 67.16 & 68.14 & 69.61 & 68.14 & 74.51 & 64.71 & 72.06 & 64.71 & 77.94 & 76.47 & 77.94 & 84.31 \\
als\_Latn & 74.51 & 74.51 & 77.94 & 78.92 & 77.45 & 79.41 & 75.00 & 68.63 & 75.98 & 75.00 & 80.88 & 83.82 & 82.84 & 84.31 \\
amh\_Ethi & 56.37 & 59.31 & 60.29 & 60.29 & 61.76 & 60.78 & 68.63 & 57.35 & 61.27 & 60.78 & 60.29 & 67.16 & 67.16 & 71.08 \\
apc\_Arab & 63.73 & 67.65 & 65.20 & 68.63 & 68.14 & 71.08 & 75.00 & 61.76 & 72.06 & 62.25 & 78.43 & 79.90 & 76.47 & 85.78 \\
arb\_Arab & 65.69 & 69.61 & 68.63 & 69.61 & 72.55 & 72.06 & 77.45 & 65.20 & 74.02 & 70.10 & 76.96 & 81.37 & 75.98 & 83.82 \\
ary\_Arab & 58.82 & 65.20 & 63.24 & 65.69 & 64.71 & 66.18 & 74.51 & 57.84 & 71.57 & 60.78 & 71.57 & 71.57 & 73.53 & 80.39 \\
arz\_Arab & 63.24 & 66.18 & 64.71 & 66.18 & 67.65 & 65.20 & 73.53 & 62.25 & 69.61 & 61.76 & 76.47 & 78.92 & 76.96 & 82.84 \\
asm\_Beng & 70.59 & 72.06 & 73.04 & 72.06 & 75.98 & 75.49 & 73.53 & 69.12 & 76.47 & 72.55 & 64.22 & 78.92 & 76.96 & 83.82 \\
ast\_Latn & 79.90 & 76.47 & 81.37 & 79.90 & 79.41 & 80.39 & 82.84 & 74.02 & 78.43 & 81.37 & 85.78 & 84.31 & 82.84 & 89.71 \\
ayr\_Latn & 39.71 & 45.59 & 43.14 & 46.08 & 44.61 & 45.59 & 42.16 & 44.12 & 47.55 & 47.06 & 46.08 & 46.57 & 53.43 & 52.45 \\
azb\_Arab & 47.55 & 50.49 & 50.98 & 50.98 & 52.45 & 54.41 & 64.71 & 48.04 & 61.27 & 49.51 & 46.57 & 60.29 & 61.76 & 70.59 \\
azj\_Latn & 77.45 & 76.47 & 77.45 & 77.94 & 78.92 & 82.35 & 77.94 & 71.57 & 79.41 & 76.47 & 78.92 & 82.84 & 83.82 & 84.80 \\
bak\_Cyrl & 69.12 & 75.49 & 71.57 & 72.55 & 72.06 & 74.51 & 75.00 & 66.18 & 73.53 & 69.61 & 77.94 & 75.00 & 78.92 & 80.88 \\
bam\_Latn & 44.12 & 46.08 & 48.04 & 47.06 & 49.51 & 50.49 & 52.94 & 43.14 & 47.55 & 44.61 & 46.08 & 50.00 & 57.35 & 49.02 \\
ban\_Latn & 75.00 & 74.51 & 76.47 & 77.94 & 78.92 & 78.43 & 79.90 & 69.61 & 73.53 & 78.43 & 75.98 & 83.33 & 84.31 & 82.35 \\
bel\_Cyrl & 76.96 & 76.96 & 75.00 & 78.43 & 77.94 & 78.92 & 76.47 & 71.08 & 77.94 & 74.02 & 76.47 & 84.80 & 81.86 & 85.78 \\
bem\_Latn & 50.49 & 52.94 & 52.45 & 49.51 & 52.45 & 54.41 & 53.43 & 47.55 & 58.82 & 50.98 & 50.49 & 54.41 & 65.69 & 62.25 \\
ben\_Beng & 72.06 & 72.06 & 67.65 & 69.61 & 73.53 & 72.55 & 79.41 & 65.20 & 75.49 & 68.14 & 66.18 & 77.94 & 77.94 & 81.86 \\
bjn\_Latn & 71.57 & 73.04 & 72.55 & 72.06 & 73.53 & 73.53 & 76.47 & 71.08 & 73.04 & 74.51 & 70.10 & 80.39 & 79.90 & 83.33 \\
bod\_Tibt & 53.92 & 50.49 & 49.51 & 50.98 & 56.86 & 50.98 & 50.98 & 48.53 & 51.96 & 51.96 & 36.76 & 56.37 & 50.98 & 59.31 \\
bos\_Latn & 78.43 & 78.43 & 75.98 & 78.92 & 81.37 & 81.86 & 79.90 & 72.06 & 82.35 & 77.94 & 80.39 & 82.84 & 82.84 & 89.22 \\
bul\_Cyrl & 77.94 & 78.92 & 77.45 & 79.41 & 78.92 & 77.94 & 78.92 & 72.06 & 78.92 & 78.43 & 81.86 & 85.78 & 80.88 & 85.78 \\
cat\_Latn & 78.43 & 77.45 & 80.88 & 83.82 & 80.39 & 83.33 & 81.86 & 74.02 & 81.86 & 79.41 & 84.31 & 86.76 & 84.31 & 88.24 \\
ceb\_Latn & 75.98 & 75.98 & 76.96 & 76.47 & 78.92 & 76.47 & 79.90 & 71.08 & 75.98 & 76.96 & 74.51 & 84.80 & 81.37 & 86.27 \\
ces\_Latn & 75.49 & 76.96 & 77.94 & 76.96 & 78.43 & 79.41 & 77.94 & 69.61 & 77.45 & 76.96 & 78.92 & 85.78 & 82.35 & 88.73 \\
cjk\_Latn & 44.12 & 44.12 & 44.12 & 44.12 & 47.06 & 46.57 & 43.14 & 40.69 & 46.08 & 49.02 & 44.61 & 50.00 & 55.88 & 47.06 \\
ckb\_Arab & 69.12 & 71.08 & 66.67 & 69.61 & 69.12 & 72.06 & 75.00 & 63.24 & 73.53 & 67.16 & 66.18 & 63.24 & 75.98 & 81.86 \\
cmn\_Hani & 76.96 & 77.45 & 81.86 & 79.90 & 79.41 & 79.90 & 86.27 & 69.12 & 81.37 & 77.94 & 82.84 & 79.90 & 80.88 & 89.71 \\
crh\_Latn & 71.08 & 69.12 & 67.65 & 72.06 & 74.02 & 72.55 & 72.55 & 62.25 & 74.02 & 69.61 & 73.04 & 75.98 & 75.49 & 75.00 \\
cym\_Latn & 77.45 & 76.96 & 75.98 & 75.98 & 80.39 & 78.43 & 77.94 & 71.57 & 78.43 & 79.90 & 70.59 & 85.78 & 84.31 & 78.92 \\
dan\_Latn & 81.37 & 82.84 & 80.88 & 82.35 & 79.90 & 83.82 & 84.80 & 75.49 & 82.35 & 82.84 & 82.35 & 86.76 & 82.84 & 89.71 \\
deu\_Latn & 80.39 & 80.88 & 83.82 & 83.82 & 82.35 & 83.33 & 83.33 & 74.51 & 78.43 & 84.31 & 85.78 & 86.27 & 85.29 & 86.76 \\
dyu\_Latn & 51.96 & 51.96 & 50.98 & 50.98 & 51.47 & 56.37 & 49.02 & 47.55 & 49.02 & 49.51 & 48.04 & 54.41 & 57.84 & 50.49 \\
dzo\_Tibt & 31.86 & 39.22 & 35.78 & 33.33 & 40.69 & 37.75 & 50.00 & 34.31 & 43.14 & 37.25 & 22.06 & 48.53 & 34.80 & 54.90 \\
ell\_Grek & 74.02 & 75.00 & 78.92 & 77.45 & 77.45 & 79.41 & 76.47 & 72.06 & 76.47 & 74.02 & 75.98 & 78.92 & 80.39 & 83.82 \\
eng\_Latn & 82.84 & 84.31 & 85.29 & 85.78 & 84.31 & 84.80 & 86.27 & 73.04 & 84.80 & 85.78 & 86.27 & 85.29 & 87.75 & 91.67 \\
epo\_Latn & 73.53 & 76.96 & 75.00 & 74.51 & 76.96 & 77.45 & 79.90 & 71.08 & 78.43 & 77.94 & 76.96 & 86.27 & 84.80 & 82.84 \\
est\_Latn & 68.63 & 73.04 & 71.08 & 71.57 & 74.02 & 75.49 & 75.00 & 67.16 & 75.98 & 71.57 & 72.55 & 79.90 & 80.39 & 79.90 \\
eus\_Latn & 59.80 & 70.10 & 69.12 & 69.61 & 69.61 & 73.04 & 75.49 & 63.73 & 75.49 & 68.63 & 69.61 & 79.90 & 86.27 & 83.82 \\
ewe\_Latn & 42.65 & 45.59 & 44.61 & 47.55 & 48.04 & 47.06 & 48.04 & 41.67 & 42.16 & 43.14 & 47.55 & 49.51 & 54.90 & 53.43 \\
fao\_Latn & 62.75 & 66.18 & 68.14 & 64.71 & 69.61 & 69.61 & 73.53 & 59.31 & 70.10 & 63.24 & 65.69 & 77.94 & 77.45 & 83.82 \\
fij\_Latn & 43.14 & 42.65 & 46.57 & 45.10 & 45.59 & 48.04 & 50.49 & 44.12 & 49.51 & 47.55 & 47.55 & 54.41 & 60.29 & 53.43 \\
fin\_Latn & 75.49 & 75.00 & 75.49 & 74.51 & 75.98 & 77.94 & 77.94 & 67.16 & 78.92 & 74.51 & 74.51 & 82.84 & 80.39 & 83.33 \\
fon\_Latn & 43.14 & 46.57 & 46.57 & 46.57 & 50.98 & 52.45 & 42.16 & 41.18 & 48.53 & 42.65 & 46.08 & 50.49 & 57.35 & 48.04 \\
fra\_Latn & 78.92 & 78.43 & 78.43 & 80.88 & 81.37 & 81.86 & 83.82 & 72.06 & 79.41 & 81.86 & 80.39 & 86.76 & 82.84 & 89.22 \\
ful\_Latn & 45.59 & 47.06 & 47.55 & 50.49 & 52.94 & 50.98 & 43.63 & 44.61 & 46.08 & 50.00 & 51.96 & 57.35 & 55.88 & 52.94 \\
fur\_Latn & 69.61 & 68.63 & 71.08 & 69.61 & 70.59 & 73.04 & 75.49 & 63.73 & 68.14 & 70.59 & 75.98 & 74.51 & 80.88 & 79.41 \\
gla\_Latn & 63.24 & 57.84 & 64.71 & 65.20 & 68.63 & 68.14 & 62.75 & 60.29 & 63.24 & 68.63 & 59.80 & 75.49 & 74.51 & 67.16 \\
gle\_Latn & 65.20 & 68.14 & 71.08 & 71.08 & 71.08 & 72.55 & 66.18 & 66.18 & 72.55 & 67.65 & 64.71 & 80.88 & 81.37 & 74.02 \\
glg\_Latn & 79.41 & 80.88 & 79.41 & 79.90 & 77.94 & 81.37 & 84.80 & 71.57 & 83.33 & 81.86 & 86.76 & 84.80 & 87.75 & 86.76 \\
grn\_Latn & 63.73 & 64.22 & 64.22 & 67.16 & 65.69 & 69.12 & 70.10 & 61.76 & 67.16 & 66.18 & 69.61 & 71.08 & 75.00 & 77.94 \\
guj\_Gujr & 73.53 & 73.04 & 70.10 & 69.61 & 77.45 & 73.04 & 79.90 & 69.61 & 73.04 & 67.65 & 62.25 & 77.45 & 78.92 & 85.29 \\
hat\_Latn & 70.10 & 72.55 & 75.00 & 73.53 & 75.00 & 76.96 & 76.47 & 65.20 & 77.45 & 73.53 & 73.04 & 82.84 & 81.37 & 82.35 \\
hau\_Latn & 68.14 & 65.69 & 67.65 & 63.73 & 67.16 & 68.14 & 69.12 & 58.82 & 68.63 & 66.67 & 61.27 & 76.47 & 74.02 & 69.12 \\
heb\_Hebr & 47.55 & 50.00 & 45.10 & 49.02 & 52.45 & 54.41 & 62.75 & 47.55 & 61.76 & 45.59 & 50.00 & 65.69 & 68.63 & 77.45 \\
hin\_Deva & 68.63 & 71.08 & 69.61 & 68.63 & 73.53 & 76.47 & 78.92 & 69.12 & 78.43 & 70.59 & 63.24 & 80.39 & 79.90 & 85.29 \\
hne\_Deva & 67.16 & 74.02 & 69.12 & 68.63 & 75.00 & 77.45 & 74.51 & 63.24 & 75.00 & 70.59 & 62.75 & 81.37 & 80.39 & 80.88 \\
hrv\_Latn & 79.41 & 80.88 & 78.92 & 77.94 & 78.92 & 82.35 & 80.39 & 73.04 & 80.88 & 76.47 & 80.39 & 84.80 & 82.35 & 86.76 \\
hun\_Latn & 76.47 & 75.00 & 77.45 & 77.45 & 77.45 & 76.96 & 80.88 & 71.08 & 76.96 & 76.96 & 77.94 & 86.27 & 78.43 & 87.25 \\
hye\_Armn & 74.02 & 75.49 & 72.06 & 74.02 & 74.51 & 75.98 & 76.47 & 66.67 & 74.51 & 71.57 & 70.59 & 79.90 & 80.39 & 84.80 \\
ibo\_Latn & 71.08 & 73.53 & 72.06 & 73.53 & 73.53 & 74.02 & 75.00 & 66.18 & 71.57 & 69.61 & 71.57 & 79.90 & 83.33 & 80.88 \\
ilo\_Latn & 66.67 & 69.12 & 71.08 & 69.61 & 73.53 & 75.49 & 74.51 & 62.75 & 71.57 & 71.57 & 74.51 & 76.47 & 83.33 & 78.92 \\
ind\_Latn & 79.41 & 81.86 & 80.88 & 80.88 & 81.37 & 82.35 & 84.80 & 74.51 & 80.88 & 83.33 & 83.33 & 84.80 & 85.78 & 90.69 \\
isl\_Latn & 69.12 & 69.12 & 69.61 & 70.59 & 71.57 & 73.04 & 74.02 & 65.20 & 68.63 & 67.65 & 69.61 & 78.92 & 72.55 & 81.37 \\
ita\_Latn & 80.88 & 79.90 & 82.35 & 82.84 & 83.33 & 83.33 & 84.80 & 73.53 & 82.35 & 83.33 & 86.27 & 87.25 & 86.27 & 90.69 \\
jav\_Latn & 72.06 & 74.02 & 72.55 & 74.51 & 75.00 & 75.49 & 77.45 & 69.61 & 75.98 & 74.02 & 76.47 & 77.45 & 78.43 & 85.78 \\
jpn\_Jpan & 78.43 & 78.43 & 80.88 & 80.88 & 83.82 & 81.86 & 83.33 & 74.51 & 77.94 & 81.86 & 79.90 & 86.76 & 80.88 & 86.27 \\
kab\_Latn & 34.31 & 36.27 & 37.75 & 37.75 & 38.73 & 39.71 & 34.31 & 33.33 & 38.24 & 33.33 & 33.82 & 40.69 & 40.20 & 35.78 \\
kac\_Latn & 40.20 & 41.67 & 41.18 & 39.71 & 45.10 & 44.12 & 42.65 & 39.71 & 40.69 & 42.65 & 47.06 & 47.06 & 54.90 & 46.08 \\
kam\_Latn & 41.18 & 43.63 & 40.69 & 44.12 & 47.55 & 47.06 & 44.12 & 40.69 & 50.98 & 44.61 & 44.61 & 51.47 & 55.39 & 49.51 \\
kan\_Knda & 69.61 & 72.06 & 66.18 & 69.61 & 71.08 & 73.53 & 74.02 & 65.20 & 74.51 & 68.63 & 64.71 & 76.96 & 77.45 & 77.45 \\
kat\_Geor & 76.47 & 75.49 & 77.45 & 74.51 & 77.45 & 77.45 & 78.92 & 74.02 & 79.41 & 76.96 & 64.71 & 83.82 & 78.92 & 83.82 \\
kaz\_Cyrl & 72.55 & 73.04 & 73.04 & 73.53 & 75.00 & 75.49 & 77.45 & 64.22 & 75.49 & 70.59 & 73.53 & 80.88 & 79.41 & 82.35 \\
kbp\_Latn & 43.14 & 44.12 & 48.04 & 48.04 & 47.55 & 48.53 & 49.51 & 43.63 & 42.16 & 46.57 & 47.55 & 44.61 & 50.98 & 47.55 \\
kea\_Latn & 74.51 & 75.98 & 75.00 & 77.94 & 75.00 & 76.96 & 79.41 & 68.14 & 77.45 & 75.49 & 82.84 & 81.37 & 82.35 & 78.92 \\
khm\_Khmr & 76.96 & 76.47 & 75.49 & 74.51 & 76.47 & 77.94 & 80.88 & 71.57 & 77.45 & 74.51 & 75.00 & 78.92 & 79.41 & 85.29 \\
kik\_Latn & 50.00 & 53.92 & 55.39 & 52.94 & 54.90 & 55.39 & 58.82 & 48.04 & 54.90 & 52.94 & 53.43 & 57.35 & 64.71 & 58.82 \\
kin\_Latn & 49.51 & 51.96 & 51.96 & 50.49 & 51.47 & 51.96 & 56.37 & 49.02 & 55.39 & 50.49 & 48.04 & 64.71 & 72.06 & 62.75 \\
kir\_Cyrl & 68.14 & 69.61 & 69.61 & 70.59 & 72.06 & 74.51 & 72.06 & 64.22 & 73.04 & 64.71 & 69.61 & 75.98 & 77.94 & 78.43 \\
kmb\_Latn & 40.20 & 41.67 & 42.16 & 43.14 & 44.61 & 46.08 & 43.63 & 43.63 & 47.06 & 44.12 & 43.14 & 48.53 & 53.43 & 50.98 \\
kmr\_Latn & 58.33 & 61.76 & 61.76 & 61.27 & 65.20 & 66.18 & 67.16 & 55.88 & 69.61 & 61.27 & 57.84 & 74.02 & 76.96 & 69.61 \\
kon\_Latn & 58.33 & 64.71 & 61.27 & 63.24 & 65.69 & 67.16 & 63.24 & 56.37 & 61.76 & 59.80 & 61.76 & 59.31 & 74.51 & 64.22 \\
kor\_Hang & 76.96 & 78.92 & 80.88 & 83.82 & 81.37 & 81.86 & 81.37 & 75.98 & 80.39 & 81.86 & 87.25 & 86.27 & 85.29 & 88.73 \\
lao\_Laoo & 72.55 & 70.59 & 74.51 & 72.06 & 73.04 & 70.59 & 75.98 & 66.67 & 73.53 & 74.02 & 70.59 & 76.47 & 78.43 & 84.31 \\
lij\_Latn & 73.04 & 73.04 & 72.06 & 74.02 & 74.51 & 75.49 & 75.98 & 67.16 & 72.55 & 73.53 & 78.43 & 76.96 & 78.92 & 80.39 \\
lim\_Latn & 77.45 & 76.47 & 75.98 & 79.41 & 77.94 & 77.45 & 75.98 & 68.14 & 78.92 & 76.96 & 81.86 & 83.82 & 82.35 & 78.43 \\
lin\_Latn & 59.31 & 57.84 & 60.78 & 63.24 & 60.29 & 59.31 & 62.75 & 52.45 & 63.24 & 60.78 & 65.20 & 61.27 & 67.65 & 66.67 \\
\bottomrule
\end{tabular}
}
\caption{7-shot accuracy with on \sib.}
\label{result1}
\end{table*}

\begin{table*}[ht]
    \centering
    \resizebox{0.8\textwidth}{!}{
\begin{tabular}{c|ccccccc|ccccccc}
\toprule
& \multicolumn{7}{c|}{Label-Aware} & \multicolumn{7}{c}{Label-Agnostic} \\
& Random & \glot & \mala & SBERT & LaBSE & Multilingual E5 & \modelname & Random & \glot & \mala & SBERT & LaBSE & Multilingual E5 & \modelname \\
\midrule
lit\_Latn & 66.67 & 68.63 & 67.65 & 69.12 & 71.57 & 68.63 & 74.02 & 66.18 & 72.55 & 70.10 & 70.10 & 78.43 & 80.39 & 86.27 \\
lmo\_Latn & 70.10 & 72.06 & 73.04 & 72.06 & 72.55 & 73.53 & 75.98 & 66.18 & 75.00 & 73.53 & 74.02 & 79.41 & 78.92 & 77.45 \\
ltz\_Latn & 74.02 & 72.55 & 74.51 & 73.53 & 77.94 & 75.49 & 76.47 & 69.61 & 78.92 & 78.43 & 76.47 & 83.82 & 83.82 & 80.88 \\
lua\_Latn & 48.53 & 48.53 & 50.00 & 49.02 & 51.96 & 50.98 & 49.51 & 48.04 & 50.49 & 50.00 & 54.41 & 57.35 & 62.25 & 53.92 \\
lug\_Latn & 43.14 & 47.55 & 49.51 & 45.10 & 47.06 & 48.04 & 46.57 & 44.61 & 49.51 & 46.57 & 45.59 & 53.43 & 61.27 & 56.37 \\
luo\_Latn & 45.10 & 46.08 & 47.55 & 48.04 & 49.51 & 51.96 & 48.04 & 44.61 & 43.63 & 47.55 & 49.51 & 54.90 & 56.37 & 58.82 \\
lus\_Latn & 55.39 & 54.41 & 56.86 & 56.37 & 59.31 & 59.31 & 58.33 & 51.47 & 53.43 & 57.84 & 57.84 & 64.71 & 63.73 & 69.12 \\
lvs\_Latn & 67.16 & 67.65 & 72.06 & 70.10 & 73.04 & 72.55 & 75.98 & 63.73 & 74.51 & 70.59 & 75.98 & 79.41 & 80.88 & 79.41 \\
mai\_Deva & 62.25 & 67.16 & 68.63 & 66.18 & 72.06 & 72.55 & 72.55 & 58.82 & 72.55 & 61.27 & 60.78 & 79.41 & 76.47 & 83.33 \\
mal\_Mlym & 65.69 & 66.18 & 66.18 & 63.73 & 68.14 & 69.12 & 75.49 & 62.25 & 69.61 & 65.20 & 54.90 & 74.02 & 75.00 & 80.39 \\
mar\_Deva & 69.12 & 72.06 & 71.57 & 72.06 & 74.51 & 74.02 & 75.49 & 66.18 & 72.55 & 70.10 & 67.65 & 79.41 & 76.96 & 80.88 \\
min\_Latn & 76.47 & 75.98 & 75.00 & 76.96 & 76.47 & 78.43 & 75.49 & 69.12 & 75.00 & 75.49 & 75.49 & 81.86 & 78.92 & 81.37 \\
mkd\_Cyrl & 74.51 & 75.49 & 77.45 & 76.47 & 76.47 & 78.43 & 77.94 & 70.10 & 78.92 & 77.94 & 77.45 & 82.35 & 81.37 & 82.35 \\
mlt\_Latn & 77.45 & 75.98 & 79.41 & 78.92 & 77.45 & 81.86 & 79.90 & 71.57 & 76.96 & 79.41 & 78.92 & 81.37 & 84.80 & 86.76 \\
mon\_Cyrl & 71.57 & 70.59 & 71.08 & 72.55 & 74.02 & 73.04 & 78.92 & 65.69 & 75.00 & 68.63 & 64.22 & 76.96 & 80.39 & 82.84 \\
mos\_Latn & 45.59 & 45.10 & 47.55 & 46.08 & 46.57 & 44.61 & 43.14 & 41.67 & 44.61 & 45.59 & 47.55 & 50.98 & 50.49 & 51.47 \\
mri\_Latn & 62.25 & 61.27 & 62.25 & 63.24 & 67.65 & 66.18 & 62.25 & 59.31 & 56.37 & 64.22 & 58.82 & 72.55 & 74.51 & 70.59 \\
mya\_Mymr & 63.73 & 59.31 & 61.27 & 64.22 & 63.24 & 67.65 & 71.08 & 59.80 & 59.80 & 54.90 & 59.80 & 70.10 & 67.16 & 76.96 \\
nld\_Latn & 80.39 & 81.37 & 81.37 & 85.29 & 84.31 & 85.29 & 84.80 & 74.51 & 83.82 & 83.82 & 84.80 & 85.78 & 86.76 & 89.22 \\
nno\_Latn & 76.47 & 76.96 & 77.94 & 80.39 & 79.41 & 79.41 & 83.33 & 74.02 & 75.49 & 78.43 & 81.86 & 85.78 & 82.35 & 90.69 \\
npi\_Deva & 72.55 & 73.04 & 70.59 & 71.57 & 75.98 & 75.49 & 77.45 & 69.61 & 78.43 & 71.57 & 63.73 & 82.35 & 81.37 & 87.25 \\
nso\_Latn & 50.98 & 53.43 & 51.47 & 56.37 & 55.88 & 57.35 & 55.88 & 48.53 & 56.37 & 50.49 & 53.92 & 59.80 & 64.71 & 59.80 \\
nya\_Latn & 54.90 & 55.39 & 58.33 & 53.92 & 60.29 & 57.84 & 64.22 & 52.45 & 62.25 & 55.88 & 54.90 & 69.61 & 71.08 & 69.61 \\
oci\_Latn & 79.41 & 77.94 & 78.43 & 80.39 & 77.94 & 80.39 & 79.41 & 68.63 & 76.47 & 78.43 & 80.88 & 83.33 & 80.88 & 85.29 \\
orm\_Latn & 38.24 & 38.24 & 40.20 & 37.75 & 41.18 & 40.69 & 36.76 & 37.25 & 39.22 & 39.71 & 37.75 & 43.14 & 55.39 & 46.08 \\
ory\_Orya & 62.75 & 65.20 & 60.78 & 57.84 & 60.78 & 64.71 & 72.55 & 59.31 & 67.65 & 58.33 & 58.33 & 69.61 & 68.14 & 80.88 \\
pag\_Latn & 67.16 & 68.63 & 68.14 & 71.08 & 72.06 & 70.10 & 75.00 & 60.29 & 71.08 & 68.63 & 75.98 & 72.55 & 76.96 & 80.39 \\
pan\_Guru & 61.27 & 67.16 & 63.73 & 64.22 & 65.20 & 65.20 & 72.06 & 59.80 & 67.65 & 62.25 & 58.82 & 73.04 & 74.51 & 78.43 \\
pap\_Latn & 74.02 & 75.49 & 76.47 & 75.49 & 75.98 & 78.43 & 78.43 & 70.59 & 75.49 & 76.96 & 78.43 & 79.90 & 84.31 & 80.39 \\
pes\_Arab & 75.00 & 75.98 & 76.47 & 75.00 & 77.45 & 75.98 & 83.33 & 73.53 & 77.45 & 75.49 & 73.53 & 82.35 & 82.35 & 87.75 \\
plt\_Latn & 57.84 & 65.20 & 60.29 & 58.82 & 62.25 & 63.73 & 55.88 & 58.82 & 64.71 & 60.29 & 58.33 & 72.55 & 75.00 & 63.73 \\
pol\_Latn & 77.94 & 80.88 & 78.43 & 81.37 & 78.92 & 79.90 & 81.86 & 74.51 & 79.41 & 80.39 & 83.82 & 82.35 & 81.37 & 87.25 \\
por\_Latn & 77.94 & 81.37 & 81.37 & 83.82 & 82.35 & 82.84 & 85.29 & 75.49 & 83.33 & 81.86 & 85.78 & 88.24 & 86.27 & 91.18 \\
prs\_Arab & 74.02 & 75.49 & 73.04 & 73.53 & 75.00 & 75.49 & 82.84 & 68.63 & 80.88 & 74.02 & 69.12 & 81.37 & 84.31 & 87.25 \\
pus\_Arab & 54.90 & 54.90 & 57.35 & 58.33 & 59.31 & 59.31 & 66.18 & 53.92 & 60.29 & 55.88 & 53.92 & 66.67 & 68.63 & 71.57 \\
quy\_Latn & 53.43 & 56.37 & 56.86 & 57.35 & 57.35 & 56.37 & 63.73 & 53.43 & 55.88 & 56.37 & 60.29 & 61.27 & 65.69 & 62.75 \\
ron\_Latn & 76.47 & 75.98 & 78.92 & 76.96 & 76.47 & 77.45 & 81.86 & 71.08 & 75.49 & 79.41 & 84.31 & 80.88 & 81.37 & 84.31 \\
run\_Latn & 48.04 & 49.02 & 52.94 & 47.55 & 52.94 & 53.92 & 54.41 & 47.06 & 48.53 & 49.51 & 50.98 & 67.16 & 70.10 & 65.69 \\
rus\_Cyrl & 79.41 & 79.90 & 78.92 & 83.82 & 81.37 & 81.86 & 82.35 & 70.59 & 80.88 & 79.90 & 85.78 & 85.78 & 85.29 & 90.20 \\
sag\_Latn & 52.94 & 50.00 & 53.43 & 52.45 & 53.43 & 52.94 & 57.84 & 49.51 & 53.92 & 49.51 & 50.49 & 51.47 & 54.90 & 59.31 \\
san\_Deva & 57.35 & 61.27 & 64.71 & 59.80 & 64.71 & 62.75 & 66.18 & 59.31 & 62.75 & 59.31 & 56.86 & 69.12 & 69.61 & 71.08 \\
scn\_Latn & 75.98 & 77.45 & 78.43 & 78.92 & 80.88 & 81.86 & 79.41 & 70.59 & 73.04 & 77.94 & 79.90 & 80.39 & 81.37 & 80.88 \\
sin\_Sinh & 69.61 & 72.55 & 66.18 & 70.10 & 72.06 & 72.55 & 73.53 & 66.18 & 70.59 & 69.12 & 67.65 & 77.94 & 75.49 & 81.86 \\
slk\_Latn & 73.53 & 75.00 & 75.49 & 75.49 & 78.43 & 77.94 & 78.92 & 69.12 & 79.90 & 75.49 & 76.47 & 83.82 & 83.33 & 84.31 \\
slv\_Latn & 74.02 & 75.49 & 75.00 & 74.02 & 77.45 & 77.45 & 75.98 & 68.63 & 73.53 & 76.47 & 74.51 & 83.33 & 80.88 & 84.80 \\
smo\_Latn & 61.27 & 62.25 & 67.16 & 65.20 & 68.14 & 69.61 & 66.67 & 60.78 & 65.69 & 64.71 & 64.22 & 73.04 & 71.57 & 73.53 \\
sna\_Latn & 54.41 & 54.90 & 55.39 & 52.94 & 55.39 & 55.39 & 50.00 & 50.00 & 54.41 & 50.98 & 51.47 & 60.29 & 70.10 & 63.24 \\
snd\_Arab & 48.04 & 50.00 & 50.00 & 53.43 & 54.41 & 57.84 & 56.37 & 50.49 & 59.31 & 50.00 & 53.92 & 63.24 & 65.69 & 66.67 \\
som\_Latn & 49.02 & 50.49 & 51.47 & 52.45 & 53.43 & 53.43 & 49.02 & 46.08 & 61.27 & 49.51 & 46.57 & 67.65 & 66.18 & 57.84 \\
sot\_Latn & 58.33 & 62.25 & 61.76 & 62.75 & 65.20 & 62.25 & 59.80 & 54.41 & 62.25 & 58.82 & 50.00 & 64.71 & 68.63 & 69.61 \\
spa\_Latn & 78.92 & 79.90 & 78.43 & 82.84 & 80.39 & 81.37 & 84.80 & 73.04 & 77.94 & 79.41 & 83.33 & 85.29 & 83.33 & 89.71 \\
srd\_Latn & 75.49 & 71.57 & 75.98 & 73.53 & 75.00 & 72.06 & 74.02 & 69.12 & 74.51 & 77.94 & 78.43 & 78.43 & 80.88 & 77.45 \\
srp\_Cyrl & 76.47 & 80.39 & 77.94 & 78.43 & 81.37 & 77.45 & 77.45 & 72.55 & 81.86 & 79.41 & 81.37 & 83.33 & 83.33 & 81.86 \\
ssw\_Latn & 50.98 & 52.45 & 56.37 & 53.43 & 55.88 & 59.31 & 60.29 & 53.43 & 58.82 & 55.39 & 49.51 & 65.20 & 69.12 & 63.73 \\
sun\_Latn & 77.94 & 75.49 & 75.98 & 79.41 & 80.39 & 79.41 & 80.88 & 74.02 & 77.45 & 76.96 & 78.92 & 83.33 & 80.88 & 86.76 \\
swe\_Latn & 75.49 & 76.47 & 76.47 & 79.41 & 79.90 & 79.41 & 83.82 & 71.08 & 76.47 & 78.92 & 75.98 & 81.86 & 82.84 & 86.76 \\
swh\_Latn & 65.20 & 62.75 & 64.71 & 63.24 & 66.18 & 65.20 & 68.63 & 61.27 & 65.69 & 60.78 & 59.80 & 73.53 & 72.06 & 77.45 \\
szl\_Latn & 72.55 & 73.04 & 75.49 & 74.02 & 75.98 & 75.49 & 76.96 & 67.65 & 71.08 & 71.57 & 75.49 & 78.92 & 81.86 & 75.00 \\
tam\_Taml & 68.14 & 67.16 & 68.63 & 68.14 & 67.65 & 69.12 & 74.02 & 62.75 & 67.16 & 68.14 & 57.35 & 75.98 & 75.98 & 78.92 \\
tat\_Cyrl & 72.06 & 75.00 & 72.55 & 72.55 & 74.51 & 75.98 & 76.47 & 67.65 & 75.98 & 73.53 & 70.59 & 82.84 & 77.45 & 81.37 \\
tel\_Telu & 66.18 & 69.61 & 66.18 & 64.22 & 72.06 & 70.10 & 75.00 & 59.31 & 73.04 & 64.22 & 65.69 & 75.98 & 75.98 & 81.86 \\
tgk\_Cyrl & 69.61 & 70.10 & 70.10 & 71.08 & 74.02 & 70.10 & 73.53 & 63.24 & 70.59 & 68.63 & 68.14 & 79.90 & 75.98 & 81.37 \\
tgl\_Latn & 76.47 & 79.41 & 79.41 & 78.92 & 78.92 & 79.90 & 81.37 & 71.57 & 78.92 & 80.39 & 74.02 & 84.80 & 84.80 & 87.75 \\
tha\_Thai & 76.96 & 78.43 & 75.49 & 76.47 & 76.47 & 78.43 & 78.43 & 70.59 & 78.43 & 72.06 & 65.69 & 79.41 & 79.90 & 85.78 \\
tir\_Ethi & 47.55 & 48.53 & 50.98 & 51.96 & 46.08 & 48.04 & 51.96 & 47.55 & 52.45 & 46.57 & 47.06 & 53.43 & 58.82 & 57.35 \\
tpi\_Latn & 76.47 & 78.43 & 79.90 & 79.41 & 79.90 & 80.39 & 85.78 & 72.55 & 77.45 & 75.00 & 78.92 & 80.88 & 82.35 & 85.29 \\
tsn\_Latn & 54.90 & 52.45 & 56.86 & 55.88 & 58.82 & 57.84 & 55.88 & 50.49 & 52.94 & 53.92 & 52.94 & 64.71 & 66.67 & 61.27 \\
tso\_Latn & 54.90 & 55.39 & 56.37 & 57.35 & 54.90 & 57.35 & 57.84 & 50.49 & 56.37 & 54.41 & 50.00 & 55.88 & 68.63 & 63.24 \\
tuk\_Latn & 61.76 & 66.18 & 65.20 & 67.16 & 70.10 & 66.18 & 69.61 & 60.78 & 72.55 & 61.27 & 66.18 & 75.49 & 75.98 & 75.98 \\
tum\_Latn & 50.00 & 53.43 & 53.43 & 50.98 & 52.94 & 55.88 & 53.43 & 50.00 & 55.88 & 49.02 & 49.51 & 62.75 & 70.59 & 67.16 \\
tur\_Latn & 77.45 & 75.98 & 76.47 & 78.92 & 75.98 & 78.43 & 81.86 & 67.65 & 79.41 & 74.51 & 84.80 & 83.33 & 79.90 & 85.29 \\
uig\_Arab & 44.12 & 49.02 & 43.63 & 45.59 & 50.00 & 53.43 & 62.25 & 45.10 & 50.98 & 42.16 & 42.65 & 58.33 & 53.43 & 66.67 \\
ukr\_Cyrl & 75.00 & 76.47 & 73.53 & 76.96 & 79.41 & 78.92 & 78.92 & 71.08 & 80.39 & 78.43 & 80.39 & 83.33 & 82.35 & 86.27 \\
umb\_Latn & 33.33 & 41.67 & 39.22 & 38.73 & 39.71 & 42.65 & 40.20 & 35.78 & 42.65 & 39.22 & 37.25 & 46.57 & 56.37 & 50.49 \\
urd\_Arab & 66.67 & 69.12 & 68.14 & 64.71 & 66.67 & 66.67 & 74.02 & 58.33 & 68.63 & 65.69 & 60.29 & 77.94 & 72.55 & 75.49 \\
uzb\_Latn & 69.61 & 70.59 & 68.14 & 69.61 & 71.57 & 71.57 & 72.55 & 65.20 & 75.49 & 67.16 & 65.69 & 81.37 & 77.45 & 77.94 \\
vec\_Latn & 77.45 & 76.96 & 78.43 & 79.41 & 80.88 & 80.39 & 78.43 & 73.53 & 76.47 & 78.43 & 81.37 & 81.86 & 80.88 & 81.37 \\
vie\_Latn & 81.37 & 79.90 & 81.86 & 79.90 & 82.84 & 82.84 & 84.80 & 74.51 & 79.41 & 82.35 & 69.61 & 85.78 & 85.78 & 87.25 \\
war\_Latn & 74.51 & 77.45 & 76.47 & 76.96 & 82.35 & 79.90 & 83.33 & 69.61 & 78.92 & 77.94 & 80.88 & 82.35 & 84.80 & 83.82 \\
wol\_Latn & 53.92 & 53.43 & 54.41 & 55.39 & 56.86 & 58.33 & 54.90 & 49.51 & 49.51 & 51.47 & 57.84 & 58.82 & 64.71 & 58.33 \\
xho\_Latn & 57.84 & 61.76 & 61.76 & 59.80 & 61.27 & 65.20 & 63.24 & 58.82 & 62.75 & 62.25 & 55.39 & 68.63 & 71.57 & 66.18 \\
yid\_Hebr & 47.06 & 48.04 & 49.51 & 53.43 & 49.02 & 54.41 & 51.47 & 47.06 & 50.00 & 49.02 & 45.59 & 57.35 & 60.78 & 62.75 \\
yor\_Latn & 47.55 & 50.49 & 51.47 & 49.51 & 52.94 & 57.35 & 50.00 & 45.59 & 48.53 & 50.00 & 51.47 & 62.25 & 62.75 & 57.84 \\
yue\_Hani & 78.43 & 81.37 & 83.82 & 82.84 & 83.82 & 86.27 & 84.31 & 71.08 & 81.86 & 80.88 & 83.82 & 85.29 & 81.86 & 85.29 \\
zsm\_Latn & 76.96 & 79.41 & 78.43 & 80.39 & 81.37 & 79.41 & 81.37 & 72.55 & 81.37 & 79.41 & 80.88 & 85.78 & 82.84 & 89.71 \\
zul\_Latn & 61.27 & 64.22 & 61.76 & 65.69 & 66.67 & 66.18 & 70.59 & 55.88 & 67.65 & 59.31 & 57.35 & 74.02 & 72.55 & 76.96 \\
\midrule
Avg & 65.24 & 66.60 & 66.75 & 67.13 & 68.51 & 69.09 & 70.18 & 61.68 & 68.55 & 66.25 & 66.59 & 73.69 & 74.61 & 75.91 \\
\bottomrule
\end{tabular}
}
\caption{7-shot accuracy with on \sib.}
\label{result2}
\end{table*}

\begin{table*}[ht]
    \centering
    \resizebox{0.8\textwidth}{!}{
\begin{tabular}{c|ccccccc|ccccccc}
\toprule
& \multicolumn{7}{c|}{Label-Aware} & \multicolumn{7}{c}{Label-Agnostic} \\
& Random & \glot & \mala & SBERT & LaBSE & Multilingual E5 & \modelname & Random & \glot & \mala & SBERT & LaBSE & Multilingual E5 & \modelname \\
\midrule
amh & 76.06 & 76.33 & 78.46 & 78.19 & 77.13 & 77.66 & 81.38 & 78.99 & 78.72 & 72.34 & 81.65 & 75.00 & 77.39 & 79.52 \\
eng & 76.05 & 82.17 & 81.96 & 81.12 & 81.22 & 81.12 & 82.81 & 75.32 & 81.43 & 81.54 & 81.86 & 83.02 & 83.02 & 86.50 \\
fra & 85.31 & 84.60 & 85.07 & 83.18 & 83.18 & 83.18 & 85.31 & 84.12 & 81.75 & 83.89 & 80.57 & 81.99 & 79.62 & 80.81 \\
hau & 63.94 & 63.20 & 62.45 & 63.01 & 63.20 & 64.31 & 65.43 & 61.15 & 61.52 & 59.48 & 62.08 & 63.94 & 64.87 & 66.91 \\
ibo & 82.40 & 85.87 & 83.73 & 85.33 & 81.60 & 85.33 & 83.73 & 83.73 & 84.80 & 85.87 & 85.60 & 82.67 & 81.87 & 78.13 \\
lin & 82.29 & 86.29 & 81.14 & 84.57 & 84.00 & 81.71 & 81.14 & 82.86 & 80.57 & 77.71 & 82.29 & 86.29 & 81.14 & 83.43 \\
lug & 70.10 & 67.16 & 67.65 & 66.18 & 65.20 & 67.16 & 66.18 & 67.16 & 66.67 & 64.71 & 66.18 & 65.20 & 64.71 & 69.12 \\
orm & 59.08 & 60.62 & 61.54 & 59.69 & 59.69 & 62.46 & 67.69 & 60.92 & 61.54 & 60.00 & 61.54 & 59.08 & 58.15 & 68.62 \\
pcm & 93.44 & 94.10 & 94.43 & 94.10 & 93.44 & 95.08 & 95.41 & 93.11 & 94.10 & 93.11 & 93.77 & 93.44 & 93.11 & 93.11 \\
run & 78.83 & 80.78 & 79.15 & 80.13 & 79.48 & 80.78 & 79.48 & 79.48 & 80.46 & 78.18 & 79.15 & 80.78 & 80.78 & 80.13 \\
sna & 69.11 & 65.58 & 66.67 & 65.85 & 62.87 & 63.96 & 62.06 & 63.69 & 63.96 & 61.25 & 63.96 & 61.79 & 65.31 & 59.08 \\
som & 62.01 & 62.37 & 64.52 & 63.08 & 62.37 & 63.80 & 63.80 & 60.57 & 62.37 & 63.44 & 61.29 & 64.87 & 63.08 & 58.42 \\
swa & 74.34 & 72.90 & 71.94 & 72.42 & 73.38 & 72.66 & 76.02 & 70.26 & 72.90 & 70.26 & 72.18 & 75.30 & 73.14 & 75.30 \\
tir & 51.84 & 55.88 & 56.25 & 58.82 & 56.25 & 57.72 & 65.44 & 57.35 & 61.03 & 54.41 & 61.76 & 61.40 & 59.56 & 65.07 \\
xho & 48.15 & 50.51 & 54.21 & 52.53 & 53.20 & 55.89 & 57.91 & 52.86 & 50.51 & 55.22 & 46.80 & 54.55 & 53.54 & 55.56 \\
yor & 84.15 & 85.30 & 85.01 & 83.57 & 84.44 & 85.30 & 86.46 & 86.74 & 85.88 & 83.86 & 84.15 & 83.29 & 82.42 & 81.84 \\
\midrule
Avg & 72.32 & 73.35 & 73.39 & 73.24 & 72.54 & 73.63 & 75.02 & 72.39 & 73.01 & 71.58 & 72.80 & 73.29 & 72.61 & 73.85 \\
\bottomrule
\end{tabular}
}
\caption{6-shot accuracy with on \masakha.}
\label{result}
\end{table*}

\end{document}